\def\CLASSINPUTtoptextmargin{0.75in}
\def\CLASSINPUTbottomtextmargin{1in}
\documentclass[conference,a4paper]{IEEEtran}
\usepackage[T1]{fontenc}
\usepackage[utf8]{inputenc}
\usepackage{cite,amsmath,amssymb,graphicx,booktabs,url,array}
\usepackage{stfloats,placeins,flafter}
\usepackage[hidelinks]{hyperref}
\usepackage{xcolor}
\newcommand{\clccodeurl}[1]{\mbox{\color{blue}\itshape\urlstyle{same}\url{#1}}}
\newif\ifanonymous
\ifdefined\AnonymousVersion\anonymoustrue\else\anonymousfalse\fi
\title{CLC-YOLO: A Compact Channel-Gated Prototype Network for Real-Time Leakage-Aware Breast Ultrasound Lesion Segmentation}
\ifanonymous
\author{}
\else
\author{\IEEEauthorblockN{1\textsuperscript{st} M. Fazri Nizar}
\IEEEauthorblockA{\textit{Department of Informatics}\\\textit{Engineering}\\
\textit{Universitas Sriwijaya}\\
Palembang, South Sumatera\\
mfazrinizar@gmail.com}
\and
\IEEEauthorblockN{2\textsuperscript{nd} Muhammad Naufal Rachmatullah}
\IEEEauthorblockA{\textit{Department of Informatics}\\\textit{Engineering}\\
\textit{Universitas Sriwijaya}\\
Palembang, South Sumatera\\
naufalrachmatullah@unsri.ac.id}
\and
\IEEEauthorblockN{3\textsuperscript{rd} Julian Supardi\textsuperscript{*}}
\IEEEauthorblockA{\textit{Department of Informatics}\\\textit{Engineering}\\
\textit{Universitas Sriwijaya}\\
Palembang, South Sumatera\\
julian@unsri.ac.id}}
\fi
\renewcommand{\footnoterule}{\kern-3pt\hrule width .4\columnwidth height .4pt\kern2.6pt}
\usepackage{eso-pic}
\AddToShipoutPictureFG{\AtTextUpperLeft{\raisebox{14pt}{\parbox[b]{\textwidth}{\footnotesize Accepted to IEEE ICAITech 2026\par\vspace{2pt}\hrule height .4pt}}}}
\ifanonymous\else\hypersetup{pdftitle={CLC-YOLO: A Compact Channel-Gated Prototype Network for Real-Time Leakage-Aware Breast Ultrasound Lesion Segmentation},pdfauthor={M. Fazri Nizar; Muhammad Naufal Rachmatullah; Julian Supardi},pdfsubject={IEEE ICAITech 2026, Paper 1571338872}}\fi
\begin{document}
\raggedbottom
\maketitle
\begingroup\renewcommand{\thefootnote}{}
\footnotetext{\textsuperscript{*}Corresponding author\par\smallskip\noindent \copyright{} 2026 IEEE. Personal use of this material is permitted. Permission from IEEE must be obtained for all other uses, in any current or future media, including reprinting/republishing this material for advertising or promotional purposes, creating new collective works, for resale or redistribution to servers or lists, or reuse of any copyrighted component of this work in other works.}
\endgroup

\begin{abstract}

Reliable breast ultrasound lesion segmentation requires accurate boundaries and evaluation that prevents patients or duplicate images from crossing data splits. We propose Channel Local Contrast (CLC), a compact refinement of the YOLO26 segmentation prototype head. CLC adds a fixed local high-pass residual controlled by 64 zero-initialized, bounded channel gates. Baseline and CLC were compared in five matched folds on each of four breast ultrasound datasets. BUS-BRA used patient-disjoint outer tests with separate inner validation. BUS-UCLM and BrEaST used patient-grouped validation folds; BUSI used duplicate-component groups because patient identifiers are unavailable. Group-macro Dice increased by 1.68, 3.11, 1.12, and 2.45 percentage points on BUS-BRA, BUS-UCLM, BUSI, and BrEaST, respectively. Only the BUS-BRA paired 95\% confidence interval excluded zero. CLC adds 64 parameters and 0.0049 giga floating-point operations (GFLOPs). At 640 pixels, single-T4, batch-one, 16-bit floating-point (FP16) TensorRT graph times were 2.1478 ms for CLC and 2.0512 ms for baseline, excluding preprocessing and postprocessing. CLC increased group-macro Dice across all four datasets with a measured T4 forward-pass overhead of 0.0966 ms. \ifanonymous Code: \href{https://anonymous.4open.science/r/CLC-YOLO-EF1E}{\textcolor{blue}{\textit{repository}}}. Rebuttal: \href{https://anonymous.4open.science/r/CLC-YOLO-EF1E/docs/REBUTTAL.md}{\textcolor{blue}{\textit{response to reviewers}}}.\else Code: \clccodeurl{https://github.com/mfazrinizar/CLC-YOLO}\fi

\end{abstract}


\section{Introduction}

Breast ultrasound (BUS) provides real-time, nonionizing imaging, but scanner settings, anatomy, and acquisition technique affect lesion appearance. Segmentation supports repeatable measurements and computer-aided analysis. BUS-BRA \cite{ref1}, BUS-UCLM \cite{ref2}, BUSI \cite{ref3}, and BrEaST \cite{ref4} provide public data with different diagnostic-verification standards. Their pixel-level reference contours are manual ultrasound annotations.

Multiple images from one patient can share anatomy, lesion appearance, and acquisition signatures. If these images cross training and test folds, a model can exploit patient-specific information, inflating measured performance. Inter-patient splitting keeps all images from a patient within one partition \cite{ref5,ref6,ref7,ref8}. BUSI also contains documented duplicate images \cite{ref9}. Patient-level partitioning has received attention in BUS analysis \cite{ref10}. Differences in grouping and evaluation protocols limit comparisons with published image-split results. A YOLO benchmark reported high within-dataset Dice under random image splitting and lower scores under cross-dataset evaluation \cite{ref11}.

U-Net \cite{ref12}, feature pyramids \cite{ref13}, and You Only Look At CoefficienTs (YOLACT) prototypes \cite{ref14} established complementary dense-prediction approaches. YOLO26 uses an end-to-end, non-maximum suppression (NMS)-free head \cite{ref15}. Recent BUS architectures use multibranch \cite{ref16} and hierarchical multiscale fusion \cite{ref17}. We investigate a small prototype refinement without replacing the detector or training objective.

Our contributions can be summarized as follows: (1) CLC, a compact prototype-head refinement combining fixed local contrast with 64 bounded, zero-initialized channel gates; and (2) matched within-dataset evaluations across four datasets with patient or duplicate grouping, sharing and conditioning ablations, group-level uncertainty, and measured central processing unit (CPU), single-T4, and training costs.

\section{Methods}

\subsection{Datasets, Annotation, and Leakage Control}

Each dataset was used for a separate training and evaluation experiment: BUS-BRA \cite{ref1}, BUS-UCLM \cite{ref2}, BUSI \cite{ref3}, and BrEaST \cite{ref4}. BUS-BRA contains 1,875 images from 1,064 patients. We retained its official patient-disjoint outer test folds and regenerated inner validation from non-test patients because views of some patients crossed the supplied train/validation assignments. BUS-UCLM contains 683 images from 38 patients, including 419 normal images. Its patient-prefix groups were kept disjoint between training and validation.

BUSI provides 780 images but no patient identifiers. Following the published duplicate audit \cite{ref9}, we linked duplicates and excluded 19 images in eight components with conflicting diagnoses, retaining 761 images in 579 groups. This proxy reduces duplicate leakage, although patient independence remains unverified. BrEaST provides one scan per patient for 256 patients. Table I summarizes diagnostic verification and manual mask provenance. Histopathology, where available, verifies the diagnosis; lesion contours were drawn on ultrasound images. Fig. 1 shows dataset examples.

\begin{table*}[!t]

\caption{Dataset and medical-evaluation integrity.}\label{tab:I}

\centering\footnotesize

\setlength{\tabcolsep}{3pt}

\begin{tabular}{p{.60in}r p{.55in}p{1.25in}p{1.25in}p{1.0in}}

\toprule[1pt]

Dataset & Images & Lesion / Normal & Evaluation Group & Diagnostic Reference & Segmentation Reference \\

\midrule[.75pt]

BUS-BRA \cite{ref1} & 1,875 & 1,875 / 0 & Inter-patient: 1,064 patient IDs, official Case folds & Biopsy-proven diagnosis (histopathology not specified) & Senior ultrasonographer mask \\

BUS-UCLM \cite{ref2} & 683 & 264 / 419 & Inter-patient: 38 patient prefixes & Biopsy-confirmed lesion diagnosis (histopathology not specified) & Radiologist contour, second review \\

BUSI \cite{ref3} & 761 & 630 / 131 & Duplicate-component proxy: 579 groups, no patient IDs & Supplied class, histology not established & Supplied manual mask \\

BrEaST \cite{ref4} & 256 & 252 / 4 & Inter-patient: 256 case IDs & Histopathologic diagnosis after biopsy (197) or \textgreater2-y follow-up (55) & Radiologist freehand mask \\

\bottomrule[1pt]

\end{tabular}

\end{table*}

\begin{figure*}[!t]

\centering

\includegraphics[width=.80\textwidth]{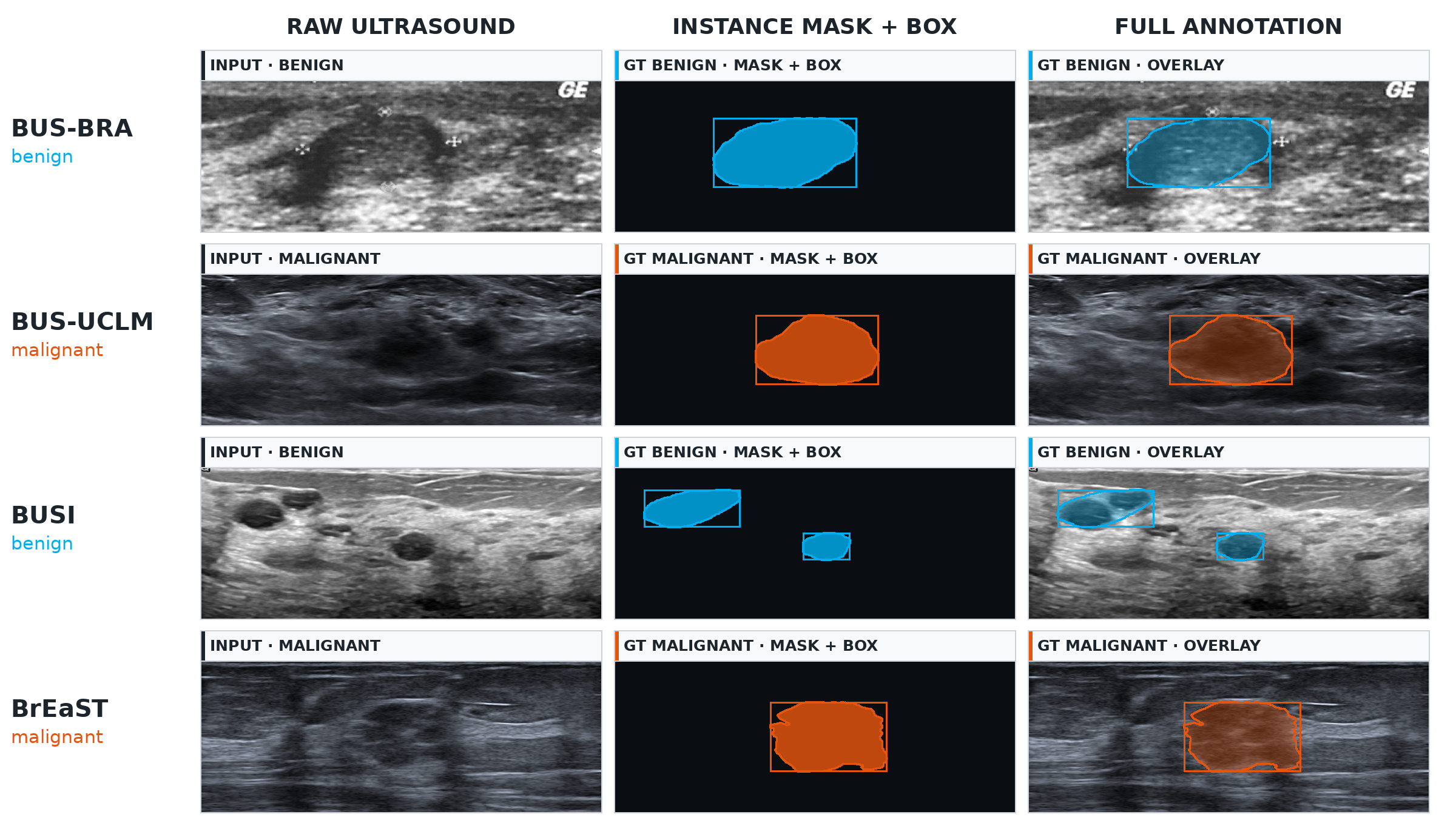}

\caption{Dataset overview: raw display crop, ground-truth (GT) instance mask and box, and annotated crop. Inference uses full images.}\label{fig:1}

\end{figure*}

Conversion retains exact masks and full image canvases. Thirteen undersized BUS-UCLM images required symmetric zero padding to match the documented canvas, a correction inferred during conversion. The image loader replicates grayscale input across three channels. Training uses aspect-ratio-preserving resize/letterboxing to 512 pixels, red/green/blue (RGB) values scaled to the unit interval, polygon targets, and mask-derived training boxes. Inference uses full images without annotation-derived cropping. Lesion-centered crops appear in the figures for display. Exact-mask evaluation uses native-resolution decoded predictions. The release includes preprocessing, polygonization, duplicate grouping, and split materialization.

\subsection{Channel Local Contrast Prototype Head}

Let \(X_3\in\mathbb{R}^{B\times64\times80\times80}\), \(X_4\in\mathbb{R}^{B\times128\times40\times40}\), and \(X_5\in\mathbb{R}^{B\times256\times20\times20}\) denote the baseline neck outputs for a 640-pixel input. Proto26 uses separate learned 1-by-1 projections \(\phi_4\) and \(\phi_5\), nearest-neighbor resize operators \(N_2\) and \(N_4\), and a 3-by-3 fusion convolution \(\psi\):

\begin{equation}\begin{gathered}F=X_3+N_2(\phi_4(X_4))+N_4(\phi_5(X_5)),\\ U=\psi(F).\end{gathered}\label{eq:1}\end{equation}

Both \(F\) and \(U\) lie in \(\mathbb{R}^{B\times64\times80\times80}\). Fusion uses elementwise addition. The auxiliary semantic branch starts from \(F\) during training, and CLC acts after \(\psi\). Let \(\Omega_{hw}\) contain the valid positions in the 3-by-3 neighborhood centered at \((h,w)\):

\begin{equation}A(U)_{bchw}=\frac{1}{|\Omega_{hw}|}\sum_{(i,j)\in\Omega_{hw}}U_{bcij}.\label{eq:2}\end{equation}

Equation (2) corresponds to average pooling with stride one and padding one, excluding padded values. CLC computes

\begin{equation}\begin{gathered}R=U-A(U),\\ \alpha=\tanh(g),\\ U'=U+\alpha\odot R.\end{gathered}\label{eq:3}\end{equation}

The new learned tensor is \(g\in\mathbb{R}^{64}\), with \(\alpha\) broadcast over both spatial axes. Positive \(\alpha_c\) enhances local variation in channel \(c\); negative \(\alpha_c\) suppresses it. Zero initialization preserves baseline behavior and gives the local-residual gradient:

\begin{equation}\begin{gathered}g=0\Rightarrow U'_{bchw}=U_{bchw},\\ \left.\frac{\partial U'_{bchw}}{\partial g_c}\right|_{g=0}=R_{bchw}.\end{gathered}\label{eq:4}\end{equation}

The unchanged generator comprises convolution (Conv) 3-by-3, transposed convolution (TransConv) 2-by-2/2, Conv 3-by-3, and Conv 1-by-1 stages \(\rho_1\) through \(\rho_4\):

\begin{equation}P=\rho_4\!\left(\rho_3\!\left(\rho_2\!\left(\rho_1(U')\right)\right)\right)\in\mathbb{R}^{B\times32\times160\times160}.\label{eq:5}\end{equation}

For retained candidate \(i\), its unchanged coefficient vector \(c_i\in\mathbb{R}^{32}\) linearly combines the shared basis. The deployed upsampled-mask path thresholds at zero and then crops with decoded box \(b_i\):

\begin{equation}\begin{gathered}L_i=c_i^{\mathsf T}P,\\ \widehat{M}_i=\operatorname{Crop}_{b_i}\!\left(\mathbf{1}\{\operatorname{Resize}(L_i)>0\}\right).\end{gathered}\label{eq:6}\end{equation}

Box, class, coefficient, loss, and postprocessing functions remain unchanged. Fig. 2 situates the head within the Path Aggregation Network--Feature Pyramid Network (PAN-FPN) neck, and Fig. 3 expands it. The added parameter count is \(\Delta\mathrm{Parameters}=|g|=64\), with a measured increase of 0.004915 GFLOPs at 640 pixels.

\begin{figure*}[t]

\centering

\includegraphics[width=\linewidth]{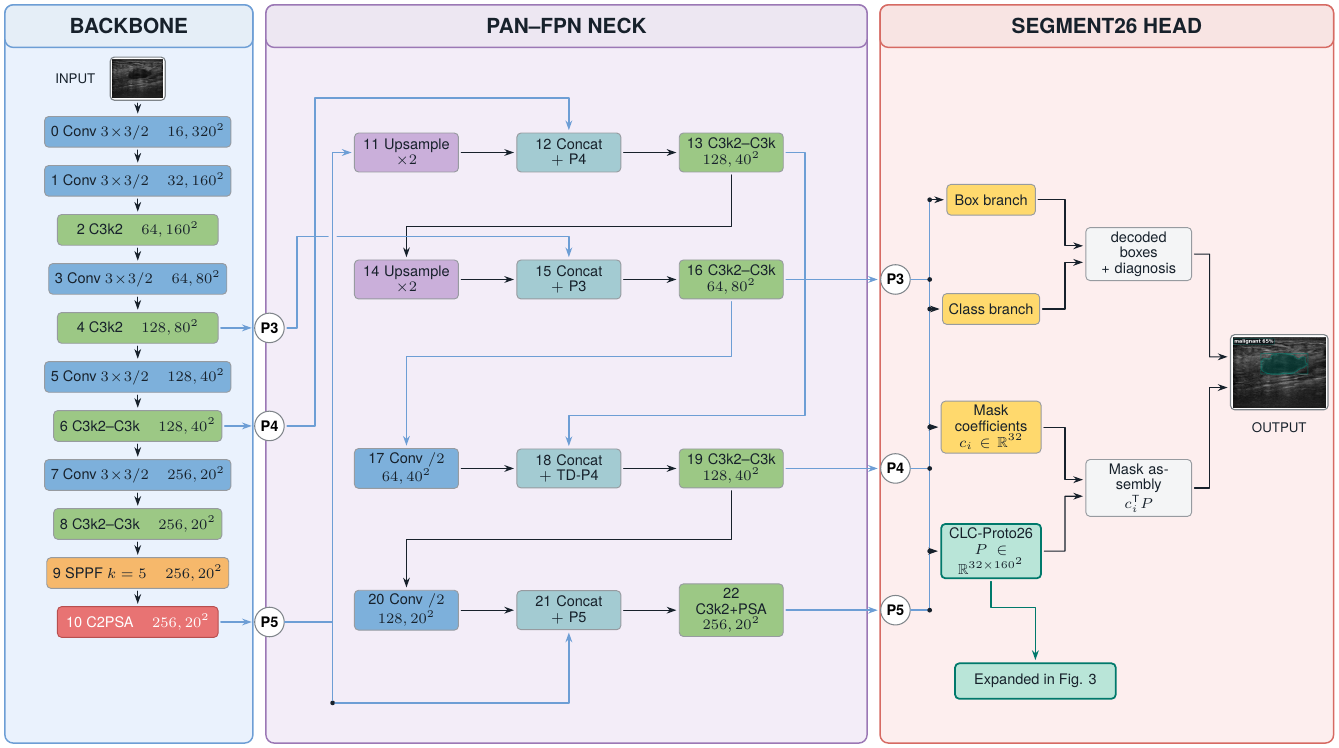}

\caption{CLC-YOLO at nano scale. Backbone and PAN-FPN neck feed Segment26; green identifies the proposed prototype refinement. Black arrows follow sequential nodes, blue arrows route lateral tensors, and breaks denote crossings without a junction. Expanded in Fig. 3.}\label{fig:2}

\end{figure*}

\begin{figure*}[t]

\centering

\includegraphics[width=\linewidth]{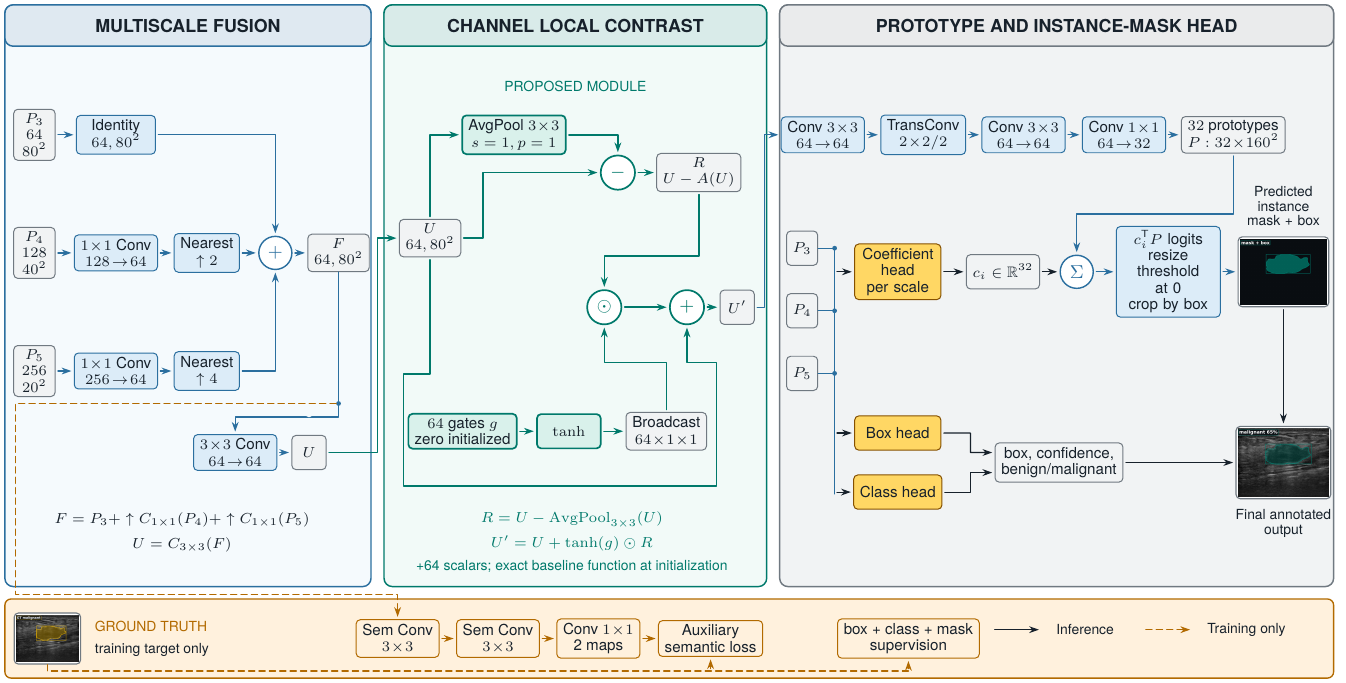}

\caption{CLC-Proto26: 64 bounded gates modulate local residuals before the unchanged prototype generator. Black/teal paths are inference; orange dashed paths are training-only. Mask coefficients combine 32 prototypes, followed by resize, threshold, and predicted-box cropping.}\label{fig:3}

\end{figure*}

Matched local-contrast (LC) controls retain \(R\) and zero functional initialization but change gate sharing or conditioning: global local contrast (Global LC) uses a single shared gate, hierarchical local contrast (HLC) combines shared and channel terms, and decoupled global-channel local contrast (DGC-LC) uses centered channel deviations. For channel logits \(v_c\), centered deviations \(d_c=v_c-\operatorname{mean}(v)\), shared logit \(a\), and mix logit \(m\), their gates are

\begin{equation}\alpha_c=\begin{cases}\tanh(a),&\text{Global LC},\\ \tanh(v_c),&\text{CLC},\\ \tanh(a+\sigma(m)d_c),&\text{HLC},\\ \tanh(a+d_c),&\text{DGC-LC}.\end{cases}\label{eq:7}\end{equation}

The initial \(\operatorname{sigmoid}(m)\) is 0.1 for HLC. For spatial mean \(\mu\) and \(\varepsilon=10^{-6}\), content-conditioned channel local contrast (C3-LC) defines \(r_{bc}=\frac{\mu(|R_{bc}|)}{\mu(|U_{bc}|)+\varepsilon}\) and \(e_{bc}=\frac{r_{bc}}{1+r_{bc}}\), followed by \(\overline{e}_b=\operatorname{mean}_c(e_{bc})\):

\begin{equation}\alpha_{bc}=\tanh\!\left(v_c+a\overline{e}_b+s_c(e_{bc}-\overline{e}_b)\right).\label{eq:8}\end{equation}

Instance-conditioned dual-basis local contrast (IDB-LC) keeps static CLC and tests instance-conditioned detail routing after prototype generation:

\begin{equation}\begin{gathered}D=P-A(P),\\ \widetilde P=[P,D],\\ \widetilde c_i=[c_i,Wc_i],\\ W=0.\end{gathered}\label{eq:9}\end{equation}

Here \(W\in\mathbb{R}^{32\times32}\). The initial logit remains \(c_i^{\mathsf T}P\); training may add \((Wc_i)^{\mathsf T}D\). These variants serve as sharing and conditioning controls. CLC-YOLO uses the static 64-gate transformation in (3).

\subsection{Training and Statistical Evaluation}

Baseline and ablations used matched folds and seeds 101, 211, 307, 401, and 503. BUS-BRA used five outer patient folds with separate grouped inner validation (about one fifth of non-test groups). BUS-UCLM, BUSI, and BrEaST used five grouped train/validation folds. For these three cohorts, checkpoint selection and reporting used the same validation fold, yielding validation estimates subject to selection optimism. Exact memberships, group counts, and per-run configurations accompany the code.

Training used Ultralytics 8.4.115, PyTorch 2.10.0, and the official COCO-pretrained YOLO26n segmentation initializer. Runs used 512 pixels, batch 64, two-T4 distributed data parallelism, automatic mixed precision, at most 100 epochs, and patience 25. The pinned automatic optimizer resolves to AdamW (initial learning rate 0.001667, momentum 0.9, weight decay 0.0005), with linear decay to 1\% and three warm-up epochs. Best checkpoints maximize native segmentation fitness. Native losses and weights remain unchanged. Matched augmentation uses horizontal flip 0.5, translation 0.1, scale 0.5, hue/saturation/value gains 0.015/0.7/0.4, and mosaic 1.0 (off in the last 10 epochs). Rotation, vertical flip, shear, perspective, mixup, cutmix, and copy-paste are zero \cite{ref18}.

The primary endpoint was class-agnostic union-mask Dice over lesion-bearing images at confidence 0.25. For group \(q\), let \(I_q^+\) contain its lesion-bearing images; \(G^+\) is the number of groups with at least one such image. With predicted union \(\widehat M_j\) and exact reference mask \(M_j\),

\begin{equation}\begin{gathered}D_j=\frac{2|\widehat M_j\cap M_j|}{|\widehat M_j|+|M_j|},\\ D_{\mathrm{group}}=\frac{1}{G^+}\sum_{q=1}^{G^+}\frac{1}{|I_q^+|}\sum_{j\in I_q^+}D_j.\end{gathered}\label{eq:10}\end{equation}

Image scores are averaged within groups, with each group given equal weight across the five reporting folds. Normal images are excluded from this Dice endpoint and assessed for false positives. Secondary endpoints are group-macro intersection over union (IoU), fold-mean mask mean average precision at IoU 0.50 (mAP50) and 0.50--0.95 (mAP50-95), and diagnosis-aware class Dice \cite{ref19}. Aggregation choices can affect biomedical model comparisons \cite{ref20,ref21}. We used 10,000 paired group-bootstrap resamples for Dice differences and 95\% confidence intervals (CIs) \cite{ref22}. These intervals quantify uncertainty in the saved predictions and remain subject to model-selection bias.

\subsection{Efficiency Measurement}

Complexity and latency measurements used 640-pixel inputs, compared with 512 pixels for training/evaluation. CPU timing used an Advanced Micro Devices (AMD) Ryzen 7 4800H, 32-bit floating-point (FP32) Open Neural Network Exchange (ONNX) Runtime 1.28.0, eight physical-core threads with fixed affinity, batch one, and both outputs. Single-T4 timing used static batch-one FP16 TensorRT 10.16.1.11 graphs, fresh weights, and nonzero surrogate CLC gates to prevent constant folding. This benchmark measures architecture cost; trained-engine accuracy remains unevaluated. Both protocols used 30 warm-up runs, 60-second timing windows, and iterative two-standard-deviation clipping. Forward-only latency excludes preparation, transfer, decoding, postprocessing, and display. Training time sums completed-epoch timestamps in the original logs, including validation and excluding setup and final evaluation.

\section{Results}

CLC increased group-macro Dice from 87.53\% to 89.21\% on BUS-BRA, 55.81\% to 58.93\% on BUS-UCLM, 69.56\% to 70.69\% on BUSI, and 67.33\% to 69.78\% on BrEaST. Changes computed before rounding were +1.68, +3.11, +1.12, and +2.45 percentage points (pp). The descriptive mean, with equal weight per dataset, rose from 70.06\% to 72.15\%.

Table II compares matched sharing and conditioning controls. CLC was the only tested variant with positive Dice changes on all four datasets. C3-LC led on BUS-UCLM and scored below baseline on BUSI and BrEaST.

\begin{table*}[!t]

\caption{Five-fold architecture ablation using patient/proxy-group macro Dice (\%). BUS-BRA is outer-test evaluation; the other cohorts are grouped validation.}\label{tab:II}

\centering\footnotesize

\setlength{\tabcolsep}{3pt}

\begin{tabular}{lrrrrrrc}

\toprule[1pt]

Variant & Added & BUS-BRA & BUS-UCLM & BUSI & BrEaST & Mean & Positive \\

\midrule[.75pt]

Baseline & 0 & 87.53 & 55.81 & 69.56 & 67.33 & 70.06 & Reference \\

Global LC & 1 & 88.86 & 52.79 & 67.03 & 63.22 & 67.97 & 1/4 \\

\textbf{Channel LC} & \textbf{64} & \textbf{89.21} & \textbf{58.93} & \textbf{70.69} & \textbf{69.78} & \textbf{72.15} & \textbf{4/4} \\

HLC & 66 & 87.68 & 52.99 & 69.07 & 65.47 & 68.80 & 1/4 \\

DGC-LC & 65 & 88.14 & 59.98 & 69.03 & 61.00 & 69.54 & 2/4 \\

C3-LC & 129 & 88.82 & 61.32 & 67.17 & 66.40 & 70.93 & 2/4 \\

IDB-LC & 1,088 & 89.00 & 57.24 & 68.59 & 62.68 & 69.38 & 2/4 \\

\bottomrule[1pt]

\end{tabular}

\end{table*}

Table III reports paired differences and fold wins. Only BUS-BRA had a paired interval excluding zero. Supplementary per-fold scores and standard deviations are supplied with the response and release; they are descriptive because folds share training data.

\begin{table*}[!t]

\caption{Paired CLC-YOLO scores (\%), with gain over baseline in parentheses (percentage points). Gains use unrounded fold aggregates. The CI and fold wins apply to Dice.}\label{tab:III}

\centering\footnotesize

\setlength{\tabcolsep}{3pt}

\begin{tabular}{lrrrrrc}

\toprule[1pt]

Dataset & Dice & IoU & mAP50 & mAP50-95 & Paired Dice 95\% CI & Fold Wins \\

\midrule[.75pt]

BUS-BRA & {89.21 (+1.68)} & {82.90 (+1.76)} & {85.41 (+1.33)} & {60.28 (+1.74)} & {{[}+0.76, +2.63{]}} & 4/5 \\

BUS-UCLM & {58.93 (+3.11)} & {53.27 (+3.13)} & {71.61 (+0.91)} & {48.92 (+1.07)} & {[}-2.70, +9.23{]} & 3/5 \\

BUSI & {70.69 (+1.12)} & {64.30 (+0.85)} & {72.28 (+1.45)} & {47.78 (+0.80)} & {[}-1.43, +3.59{]} & 3/5 \\

BrEaST & {69.78 (+2.45)} & {62.67 (+2.54)} & {70.64 (+1.57)} & {44.40 (+1.93)} & {[}-2.43, +7.43{]} & 4/5 \\

\bottomrule[1pt]

\end{tabular}

\end{table*}

CLC improved benign and malignant Dice on BUS-BRA, BUS-UCLM, and BrEaST, while BUSI malignant Dice changed from 56.87\% to 56.42\%. Normal-image false positives changed from 19/419 to 20/419 on BUS-UCLM, 12/131 to 17/131 on BUSI, and 0/4 to 2/4 on BrEaST. Only four BrEaST normals make its count descriptive. Table IV summarizes efficiency.

\begin{table*}[!t]

\caption{Complexity, inference latency, and training efficiency.}\label{tab:IV}

\centering\footnotesize

\setlength{\tabcolsep}{3pt}

\begin{tabular}{l r r p{.93in}p{.93in}p{.85in}r}

\toprule[1pt]

Model & Train Parameters & GFLOPs & Ryzen CPU ONNX (ms) & T4 TensorRT 10 (ms) & Total Training (h) & Weighted s/Epoch \\

\midrule[.75pt]

Baseline & 3,053,510 & 8.957952 & 33.2815 ± 0.7092 & 2.0512 ± 0.0102 & 4.3861 (1,901 epochs) & 8.3062 \\

\textbf{CLC-YOLO} & \textbf{3,053,574} & \textbf{8.962867} & \textbf{34.8376 ± 1.0644} & \textbf{2.1478 ± 0.0098} & \textbf{4.5104 (1,964 epochs)} & \textbf{8.2675} \\

Difference & +64 (+0.0021\%) & +0.004915 (+0.0549\%) & +1.5561 (+4.68\%) & +0.0966 (+4.71\%) & +0.1243 (+2.83\%) & -0.0387 (-0.47\%) \\

\bottomrule[1pt]

\end{tabular}

\end{table*}

CLC added 64 parameters (+0.0021\%) and 0.004915 GFLOPs (+0.0549\%). Forward latency increased by 1.5561 ms on CPU and 0.0966 ms on T4. Weighted seconds per epoch decreased by 0.47\%; cumulative time rose 2.83\% because CLC completed 63 more epochs. Fig. 4 illustrates selected cases with favorable paired Dice differences.

\begin{figure*}[!t]

\centering

\includegraphics[width=.80\textwidth]{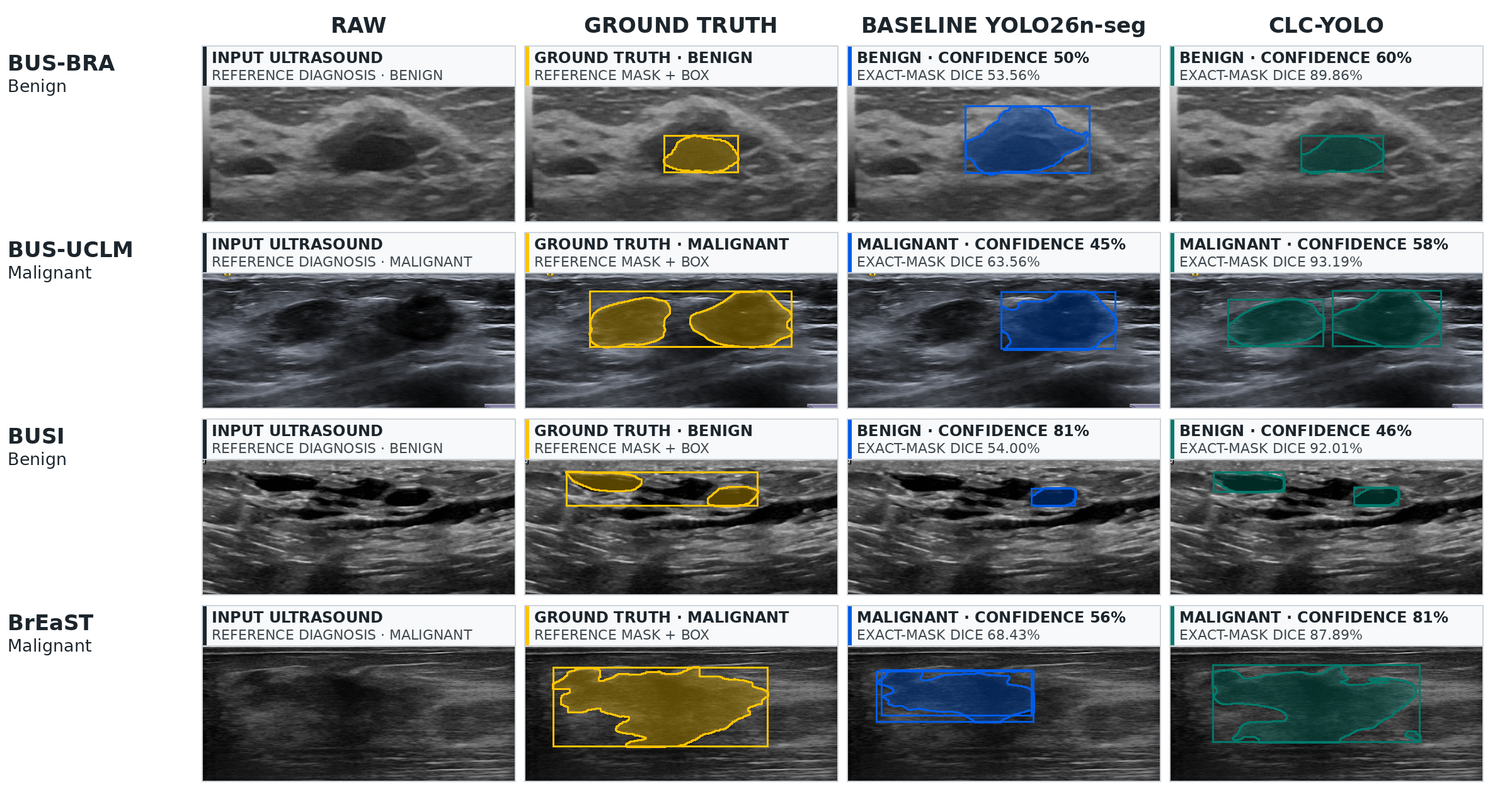}

\caption{Two benign and two malignant cases selected for favorable CLC Dice differences, with nonempty baseline masks. Columns: raw image, GT, baseline, CLC. Selection favors visible improvements over typical performance.}\label{fig:4}

\end{figure*}

\section{Discussion}

A shared gate applies the same multiplier to all 64 prototype channels. Independent bounded gates allow channel-specific enhancement or suppression of local variation. Existing designs include channel gating \cite{ref23}, convolutional block attention module (CBAM) channel-spatial attention \cite{ref24}, coordinate attention \cite{ref25}, and efficient channel attention \cite{ref26}. CLC applies bounded channel gates to a fixed high-pass residual within the YOLO26 prototype path. Zero initialization preserves baseline behavior, and the unchanged loss isolates the feature-path modification.

The ablations compare gate sharing in HLC and DGC-LC, image conditioning in C3-LC, and instance-conditioned detail in IDB-LC. CLC achieved the highest four-dataset mean. Architecture selection reused these folds, leaving selection bias in the reported gains. Independent evaluation is needed to confirm the selected design.

ONNX and TensorRT latency increased by about 4.7\%, alongside a 0.0549\% increase in operation count. Memory traffic, extra operator launches, and limited fusion may explain the difference, although their contributions were not profiled. The 2.1478-ms T4 graph time supports real-time forward computation. End-to-end clinical latency and low-power edge performance remain untested.

Higher aggregate overlap accompanied a decline in BUSI malignant Dice and more normal-image false positives. Ramadini and Supardi improved peak signal-to-noise ratio in brain magnetic resonance super-resolution without improving structural similarity \cite{ref28}. For CLC, threshold calibration and boundary-distance evaluation remain open tasks. Small group counts limit precision \cite{ref27}. The three non-BUS-BRA intervals include zero, and their reported scores reuse validation for checkpoint selection. BUSI patient independence remains unverified. Different protocols in larger-model studies \cite{ref11,ref16,ref17} limit comparison. External generalization, clinical utility, and performance against other segmentation families require further evaluation.

\section{Conclusion}

CLC-YOLO refines 64 fused prototype channels with zero-initialized local-contrast gates, adding 64 parameters and 0.0049 GFLOPs. The BUS-BRA Dice gain had a positive paired interval. The three grouped-validation cohorts had positive point estimates with intervals spanning zero. Single-T4 graph timing was 2.15 ms, with an overhead of 0.0966 ms over baseline. Independent external testing, threshold calibration, and end-to-end timing are needed before clinical deployment.

\section*{Data Availability Statement}

\ifanonymous Source code, trained weights, and leakage-aware split manifests are publicly available at \href{https://anonymous.4open.science/r/CLC-YOLO-EF1E}{\textcolor{blue}{\textit{the anonymous repository}}}.\else Source code, trained weights, and leakage-aware split manifests are publicly available at \clccodeurl{https://github.com/mfazrinizar/CLC-YOLO}.\fi\space Source datasets are available through their original papers \cite{ref1,ref2,ref3,ref4}.

\FloatBarrier
\bibliographystyle{IEEEtran}
\bibliography{references}

\end{document}